\documentclass{article}
\usepackage{amsmath,amsfonts}
\usepackage{graphicx}
\usepackage{array}
\usepackage{multirow}
\usepackage{tabularx}

\usepackage{textcomp}
\usepackage{stfloats}
\usepackage{url}
\usepackage{verbatim}

\usepackage{color}
\definecolor{purple}{rgb}{0.6,0,0.6}

\usepackage[colorlinks=true, allcolors=blue]{hyperref}
\usepackage{bm}
\usepackage{amssymb}
\usepackage{booktabs}
\usepackage{multirow}
\usepackage{algorithm,algorithmic}

\def\BibTeX{{\rm B\kern-.05em{\sc i\kern-.025em b}\kern-.08em
    T\kern-.1667em\lower.7ex\hbox{E}\kern-.125emX}}
\usepackage{balance}
\usepackage{amsthm}

\usepackage{epstopdf}
\usepackage{amsthm}

\usepackage{lineno}
\usepackage{xcolor}
\usepackage{cite}
\usepackage{caption}
\usepackage{subcaption}

\usepackage{comment}
\usepackage{amssymb}
\newtheorem{definition}{Definition}

\DeclareMathOperator*{\argmin}{arg\,min}

\begin{document}

\title{Multi-view MERA Subspace Clustering}

\author{Zhen Long$^{*}$, Ce Zhu$^{*}$, Jie Chen$^{*}$, Zihan Li$^{*}$, Yazhou Ren$^{\dag}$,Yipeng Liu
 \thanks{
 Z. Long, C. Zhu, J. Chen, Z. Li and Y. Liu are with the School of Communication and Information Engineering, and University of Electronic Science and Technology
of China (UESTC), Chengdu
611731, China. (email: eczhu@uestc.edu.cn; yipengliu@uestc.edu.cn)
\quad
$^{\dag}$Y. Ren is with the School of Computer Science and Engineering, UESTC, Chengdu, 611731, China.}
}



\maketitle

\begin{abstract}
Tensor-based multi-view subspace clustering (MSC) can capture high-order correlation in the self-representation tensor. Current tensor decompositions for MSC suffer from highly unbalanced unfolding matrices or rotation sensitivity, failing to fully explore inter/intra-view information. Using the advanced tensor network, namely, multi-scale entanglement renormalization ansatz (MERA), we propose a low-rank MERA based MSC (MERA-MSC) algorithm, where MERA factorizes a tensor into contractions of one top core factor and the rest orthogonal/semi-orthogonal factors. Benefiting from multiple interactions among orthogonal/semi-orthogonal (low-rank) factors, the low-rank MERA has a strong representation power to capture the complex inter/intra-view information in the self-representation tensor. The alternating direction method of multipliers is adopted to solve the optimization model. Experimental results on five multi-view datasets demonstrate MERA-MSC has superiority against the compared algorithms on six evaluation metrics. 
Furthermore, we extend MERA-MSC by incorporating anchor learning to develop a scalable low-rank MERA based multi-view clustering method (sMREA-MVC). The effectiveness and efficiency of sMERA-MVC have been validated on three large-scale multi-view datasets.
To our knowledge, this is the first work to introduce MERA to the multi-view clustering topic. The codes of MERA-MSC and sMERA-MVC are publicly available at https://github.com/longzhen520/MERA-MSC.
\end{abstract}

\section{Introduction}
Multi-view data are ubiquitous in the real world. For instance, certain news can be reported in text, images, and video; the multi-view Yale database
can be represented by local binary patterns (LBP), Gabor, and intensity features. Such multi-view data, which provide consensual and complementary information, have given rise to a series of multi-view learning based tasks~\cite{zheng2015closed,tang2018learning,xu2023adaptive}. Among them, multi-view subspace clustering (MSC) separates multi-view data into several groups by assuming it shares a common latent low-dimensional subspace, driving numerous applications in image processing and computer vision~\cite{xia2021self, wang2017exclusivity, shu2022self, xiao2020reliable,ren2020deep}.

Inspired by the success of low-rank representation (LRR)~\cite{vidal2014low} and sparse subspace clustering (SSC)~\cite{elhamifar2013sparse} on single-view data, where each sample can be represented by a combination of other  samples, many self-representation based MSC methods have been proposed. Among them, tensor-based MSC ones show promising performances by assuming that their self-representation tensors are low-rank~\cite{zhang2015low, xie2018unifying, fu2022unified, chen2021self}. For example, Zhang et al. \cite{zhang2015low} firstly stack all subspace representations of each view to a 3rd-order tensor and consider Tucker low-rank constraint~\cite{liu2012tensor} on it to explore the high-order correlations among views. However, Tucker nuclear norm is denoted as the sum of the nuclear norms of the unfolding matrices, which captures the low-rank information from unbalanced unfolding matrices, resulting in inferior performance~\cite{xie2018unifying, bengua2017efficient}. 
To alleviate it, Xie et al.~\cite{xie2018unifying} consider the tubal norm based on tensor singular value decomposition (t-SVD) \cite{kilmer2011factorization} for low-rank self-representation tensor, where the tensor is rotated to effectively capture the consensus among multiple views.

However, the self-representation tensor contains both inter- and intra-view similarity information. For t-SVD, the rotation operation of the self-representation tensor can well explore the correlations across different views, but it also suffers from inadequate exploration of intra-view information, because matrix singular value decompositions (SVDs) are only performed in the first two modes, and linear transformations are considered in the third mode \cite{liu2022tensor, liu2022tensors, liu2018improved }. On account of these, Jia et al.~\cite{jia2021multi} introduce low-rankness to the frontal and the horizontal slices of the self-representation tensor to characterize the intra-view and inter-view relationships, respectively.
On exploring inter/intra-view information in the self-representation tensor, one question naturally arises: whether the inter/intra-view information can be well captured simultaneously with a new tensor decomposition. In fact, as advanced tensor decompositions, tensor networks show significant advantages in capturing the correlation of higher order data~\cite{bengua2017efficient, wang2017efficient, long2021bayesian}.  

\begin{figure*}[t!]
 \centering
\includegraphics[width=0.95\textwidth]{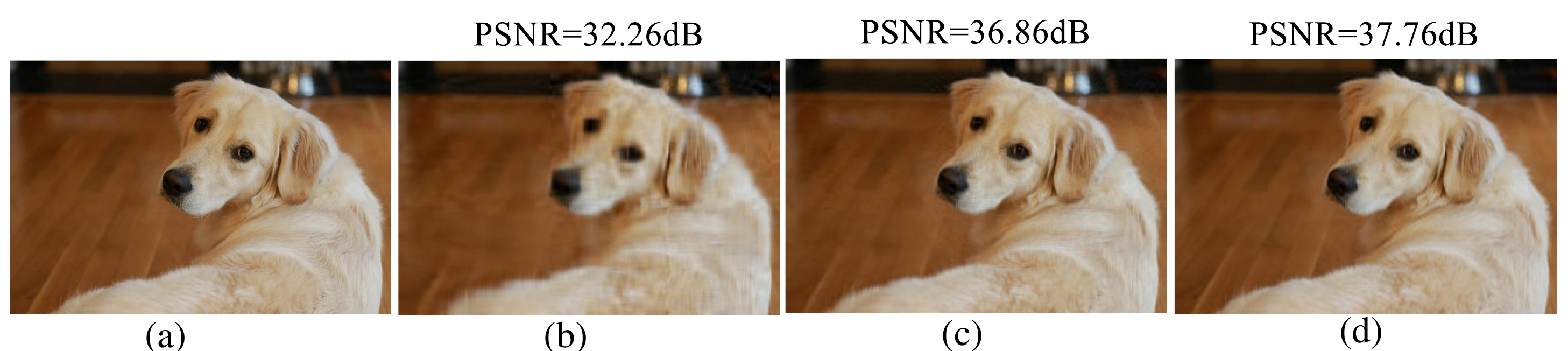}
\caption{Low-rank exploration performance of t-SVD, Tucker, and MERA on RGB image at the same 5\% compression ratio; (a) is the original image with size $640\times 960\times 3$; The images in (b)-(d) correspond to the recovered images obtained by low-rank t-SVD approximation, low-rank Tucker approximation, and low-rank MERA approximation, respectively.}
\label{fig:sparseMERA}
 \end{figure*}

In this paper, we consider a stable tensor network, namely the multi-scale entanglement renormalization ansatz (MERA) decomposition~\cite{vidal2008class,batselier2021MERAcle,ye2018tensor}, to capture well both inter- and intra-view information of the self-representation tensor. MERA factorizes a $D$-th order tensor into several orthogonal/semi-orthogonal factors and one top core factor, providing a more expressive and powerful representation of a high-order tensor. 
 To make it clear, Fig. \ref{fig:sparseMERA} provides the reconstructed images obtained by different tensor decompositions, including t-SVD, Tucker decomposition, and MERA decomposition, under the same compression ratio (CR), i.e., CR=5\%. From Fig. \ref{fig:sparseMERA}(b-d), it is evident that the low-rank MERA approximation exhibits better recovery performance in terms of PSNR than the other approximation methods. This implies that MERA has a more powerful representation capability in comparison to t-SVD and Tucker decomposition.


 \begin{figure*}[t!]
 \centering
\includegraphics[width=1\textwidth]{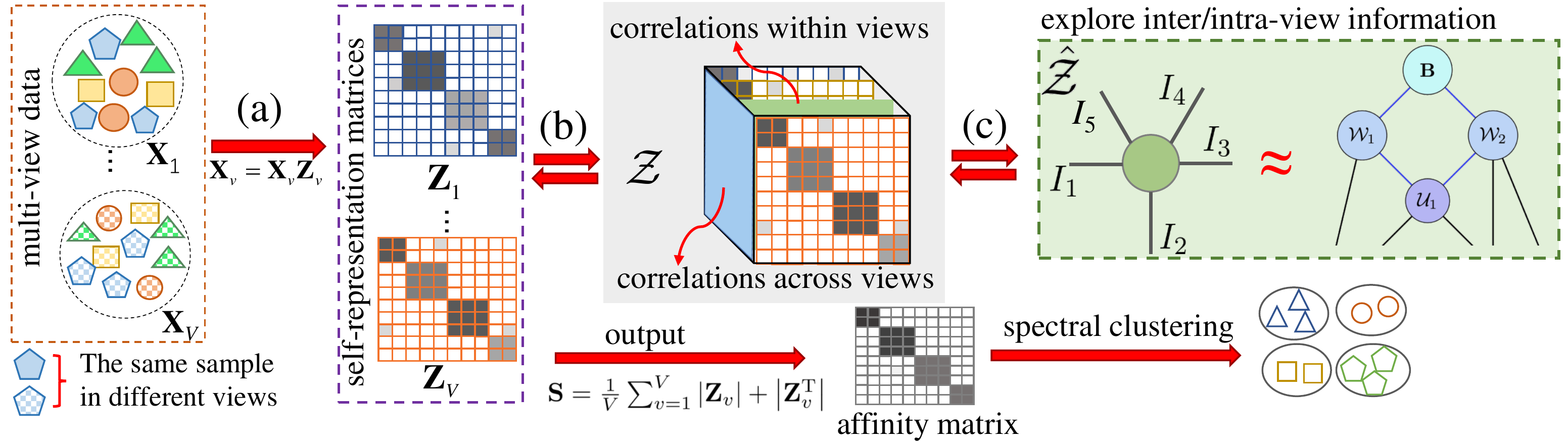}
\caption{Framework of MERA-MSC. Each data $\mathbf{X}_v$ can be first expressed with its linear combination, e.g., (a) $\mathbf{X}_v=\mathbf{X}_v\mathbf{Z}_v$; (b) these self-representation matrices $\mathbf{Z}_v,v=1,\cdots, V$ are stacked into a self-representation tensor $\mathcal{Z}$; (c) $\mathcal{Z}$ will be reshaped into a 5-th order tensor, on which low-rank MERA approximation is performed to capture both inter- and intra-view correlations. In each iteration, $\mathbf{Z}_v$ is adaptively obtained from both $\mathbf{X}_v$ and the updated $\hat{\mathbf{Z}}_v$ from low-rank MERA approximation. At last, the affinity matrix can be constructed by $\frac{1}{V}\sum_{v=1}^{V}|\mathbf{Z}_v|+|\mathbf{Z}_v^{\operatorname{T}}|$, which will then be fed into the spectral clustering algorithm.}
\label{fig:framework1}
\vspace{-0.2cm}
 \end{figure*}



 Based on it, we develop a low-rank MERA-based MSC (MERA-MSC) optimization model, and the alternating direction method of multipliers (ADMM) framework~\cite{boyd2011distributed} is adopted to solve it. The processing of MERA-MSC is illustrated in Fig. \ref{fig:framework1}, where the self-representation tensor will be adaptively updated from the multi-view data and low-rank MERA approximation in each iteration.
 Iteratively, the final self-representation tensor will be used to construct the affinity matrix, which will then be fed into the spectral clustering algorithm\cite{von2007tutorial} for achieving MSC task.
 Furthermore, we extend the MERA-MSC method using anchor learning and developed a scalable MERA based multi-view clustering (sMERA-MVC) algorithm in the experimental part. 

Our contributions are summarized as follows:
\begin{itemize}
    \item  Motivated by the powerful representation ability of MERA decomposition, we apply it to capture the inter/intra-view correlations within the self-representation tensor for MSC. To the best of our knowledge, this is the first work to introduce MERA to MVC tasks.
    \item An effective algorithm named MERA-MSC is proposed and evaluated on five benchmark datasets. Experimental results show MERA-MSC outperforms the state-of-art methods, as evaluated using six commonly used metrics. Building upon this, an sMERA-MVC is further developed, which clusters large-scale multi-view data effectively and efficiently.

    
   \item  The outstanding clustering performance of MERA-MSC and sMERA-MVC highlights the capability of low-rank MERA approximation to explore higher-order correlation, which  also provides some inspiration for applying it to other tasks, like tensor classification, tensor regression.
    %

  %


\end{itemize}





\begin{figure}
 \centering
\includegraphics[width=0.65\textwidth]{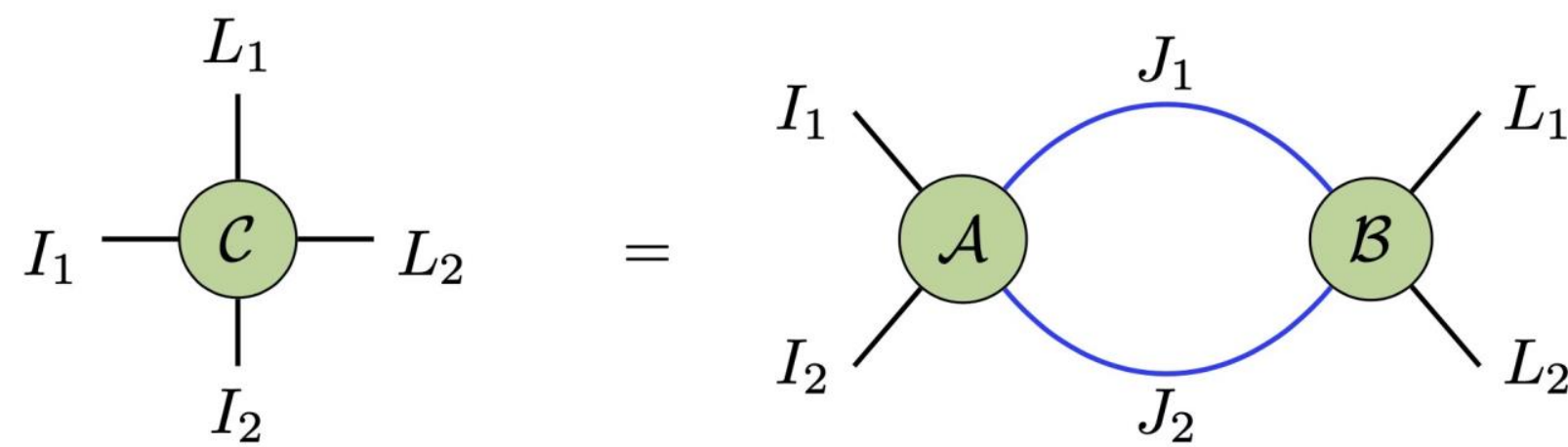}
\caption{The graphical illustration of tensor contraction.}
\label{fig:contraction}
\vspace{-0.5cm}
 \end{figure}
\section{NOTATIONS AND PROBLEM FORMULATION}

\subsection{ Notations}
A scalar, a vector, a matrix, and a tensor are written as $x$, $ \mathbf{x} $,  $  \mathbf{X} $, and $\mathcal{X}$, respectively.
Indices typically range from 1 to their capital version, e.g., $i = 1,\cdots, I$.
For a $D$-th order tensor $\mathcal{X}\in\mathbb{R}^{I_1\times \cdots \times I_D}$, its mode-\{$d_1,d_2$\} unfolding matrix is represented as $\mathbf{X}_{\left(d_{1}, d_{2}\right)} \in \mathbb{R}^{I_{d_{1}} I_{d_{2}} \times \prod_{e \neq d_{1}, d_{2}} I_{e}}$ by arranging the \{$d_1,d_2$\}-th mode of $\mathcal{X}$ as the row while the rest modes as the column, and its inverse operator is defined by $\mathcal{X}$=$\operatorname{fold}_{d_{1}, d_{2}}(\mathbf{X}_{\left(d_{1}, d_{2}\right)})$.
As shown in Fig. \ref{fig:contraction}, the tensor contraction of two tensors $\mathcal{A}\in \mathbb{R}^{I_{1}\times I_{2}\times J_{1}\times J_{2}}$ and $\mathcal{B}\in\mathbb{R}^{J_{1}\times J_{2}\times L_{1}\times L_{2}}$ is achieved by contracting their common indices $\{J_{1},J_{2}\}$, which is denoted as  $\mathcal{C}= \mathcal{A}\times_{\{J_{1},J_{2}\}}\mathcal{B}\in\mathbb{R}^{I_{1}\times I_{2}\times L_{1}\times L_{2}}$ whose entries are calculated by
$c_{i_{1},i_{2},l_{1},l_{2}}=\sum_{j_{1},j_{2}}a_{i_{1},i_{2},j_{1},,j_{2}}b_{j_{1},j_{2},l_{1},l_{2}}.$

\vspace{-0.5cm}
\subsection{ MERA Decomposition}

\begin{figure}
 \centering
\includegraphics[width=0.65\textwidth]{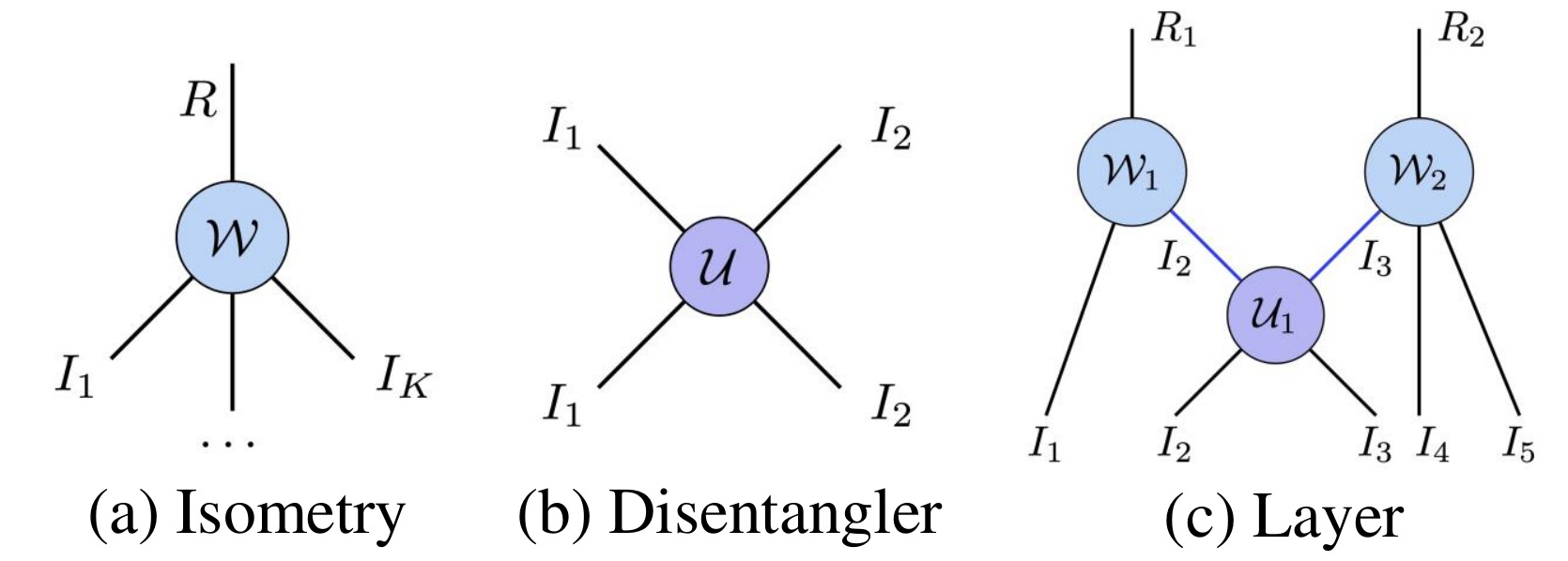}
\caption{ The graphical representations of isometry, disentangler, and layer, respectively.}
\label{fig:blocks}
\vspace{-0.5cm}
 \end{figure}

Before introducing the MERA decomposition, it is necessary to introduce two fundamental building blocks: isometry and disentangler.
 \begin{definition} [Isometry]\cite{vidal2008class,batselier2021MERAcle}
As shown in Fig. \ref{fig:blocks}(a), an isometry $\mathcal{W}\in\mathbb{R}^{I_1\times\cdots \times I_K\times R}$ is denoted as a $(K+1)$-th order tensor with semi-orthogonal unfolding matrix $\mathbf{W}\in\mathbb{R}^{I_1\cdots I_K\times R}$, e.g., $\mathbf{W}^{\operatorname{T}}\mathbf{W}=\mathbf{I}_{R}$,
where $R$ is the outgoing dimension of an isometry and $K$ is the number of bottom legs. 
\end{definition}

\begin{definition} [Disentangler]\cite{vidal2008class,batselier2021MERAcle}
As shown in Fig. \ref{fig:blocks}(b), a disentangler $\mathcal{U}\in\mathbb{R}^{I_1\times I_2 \times I_1 \times I_2}$ is denoted as a 4-way tensor with a square orthogonal unfolding matrix $\mathbf{U}\in\mathbb{R}^{I_1 I_2 \times I_1  I_2}$, namely, $\mathbf{U}^{\operatorname{T}}\mathbf{U}=\mathbf{I}_{I_1I_2}$.
\end{definition}
\begin{definition} [Layer]\cite{vidal2008class,batselier2021MERAcle}
A layer $\mathcal{C}$ is denoted as a result of tensor contraction between isometries and disentanglers in the same layer. For example, as shown in Fig. \ref{fig:blocks}(c), a 7-th order layer $\mathcal{C}$ can be constructed by two isometries $\mathcal{W}_1\in\mathbb{R}^{{I}_1\times {I}_2 \times R_1}$, $\mathcal{W}_2\in\mathbb{R}^{{I}_3\times {I}_4 \times {I}_5\times R_2}$ and one disentangler $\mathcal{U}\in\mathbb{R}^{I_2\times I_3\times I_2\times I_3}$, as follows:
\begin{equation*}
\mathcal{C}=\mathcal{U}_{1}\times_{\{I_2\}}\mathcal{W}_{1}\times_{\{I_3\}}\mathcal{W}_2\in\mathbb{R}^{I_1\times I_2 \times I_3 \times I_4 \times I_5 \times R_1\times R_2 }.
\end{equation*}
\end{definition}
 
\begin{figure}
 \centering
\includegraphics[width=0.65\textwidth]{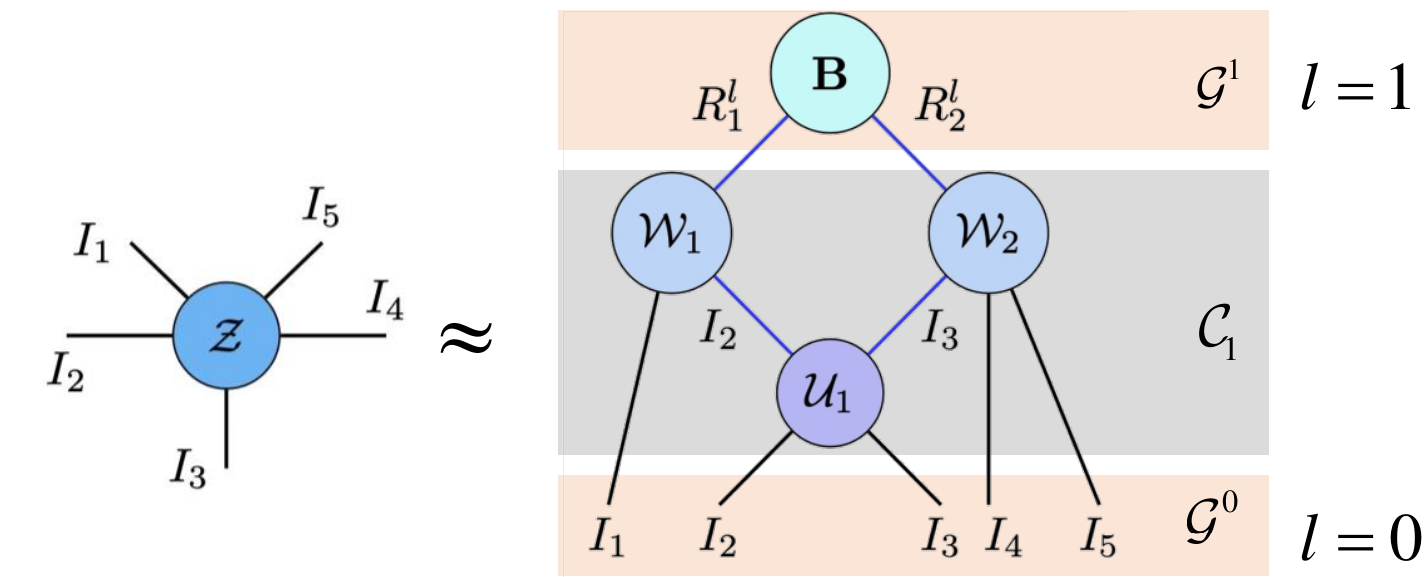}
\caption{ An graphical illustration of MERA decomposition for a 5-th order tensor.}
\label{fig:MERA}
\vspace{-0.5cm}
 \end{figure}

\begin{definition}[MERA Decomposition]\cite{vidal2008class,batselier2021MERAcle}
    For a $D$-th order tensor $\mathcal{Z}\in\mathbb{R}^{I_1\times I_2\times\cdots\times I_D}$, its  MERA representation can be composed by different layers and one top core, which can be recursively obtained by:
  \begin{equation}
      \mathcal{G}^{l-1}=\mathcal{G}^{l}\times_{\{R^{l}_1,\cdots, R^{l}_{P_l}\}}\mathcal{C}_{l},
  \end{equation}
   where $R^{l}_{p_l}, p_l=1,\cdots, P_l, l=1,\cdots,L$ are the MERA ranks; $P_l$ is the number of isometries in $l$-th layer, and $L$ is the total number of layers.
   $\mathcal{G}^0=\mathcal{Z}$, and $\mathcal{G}^{L}=\mathbf{B}$, where $\mathbf{B}$ is the top core.  In a nested way, we can construct the MERA decomposition.
\end{definition}
  For simplification, we use $\mathcal{Z}=f(\mathbf{B},\mathcal{W},\mathcal{U})$ to represent an MERA decomposition, where $\mathcal{U}=(\mathcal{U}_{1}^l,\cdots,\mathcal{U}_{P_l-1}^l) $ and $\mathcal{W}=(\mathcal{W}_{1}^l,\cdots,\mathcal{W}_{P_l}^l)$, $\mathcal{W}^l_{p_l}$ and $\mathcal{U}^l_{p_l}$ are the $p_l$-th isometry and disentangler in the $l$-th layer, $l=1,\cdots,L$.
  Fig. \ref{fig:MERA} shows the graphical representation of  MERA decomposition for a 5-th order tensor.
 
\subsection{Related Works}
According to different representation forms of self-representation matrices in multi-view data,
current MSC methods are mainly divided into two categories: one is matrix-based MSC, which learns a shared self-representation matrix of all views; the other is tensor-based MSC, which stacks 
self-representation matrices of all views into a tensor to explore the correlations among  views.
\subsubsection{Matrix-based MSC}
Current matrix-based MSC methods generally consider that multi-view data possess a shared consistent feature while each view has its specific feature that provides complementary information, namely consistency and diversity~\cite{wang2017exclusivity, li2021consensus}.
 Specifically,~\cite{wang2017exclusivity} considers a shared indicator matrix and a position-aware exclusivity term as the consistent and the diversity representations, respectively. \cite{luo2018consistent} divides the self-representation matrix of each view into two parts: the low-rank consensus shared by all views and the specific part of every single view constrained by $\ell_{2}$ norm.
 Furthermore, instead of the original space,~\cite{zhou2019dual} projects multi-view data into a latent space to exploit shared-specific information.
Besides,~\cite{si2022consistent} considers the interaction between the learned self-representation and the clustering calculation and proposes a unified MSC framework to learn a better subspace self-representation for clustering.
\subsubsection{Tensor based MSC}
Tensor-based MSC methods can well capture high-order correlations among multiple views, showing improved clustering performance~\cite{zhang2015low, tang2021constrained}. The key idea of tensor-based MSC methods is to find a good low-rank tensor approximation for the self-representation tensor.
Given multi-view dataset $\mathbf{X}_v\in\mathbb{R}^{D_v\times N}$, $v=1,\cdots, V$, where $V$ and  $N$ are the numbers of views and samples, respectively, and $D_v$ represents the feature dimension in the $v$-th view. The general tensor-based MSC optimization model can be formulated as~\cite{zhang2015low}:
\begin{eqnarray}\label{general model}
   &&\min_{\{\mathbf{E}_v,\mathbf{Z}_v\}_{v=1}^{V}} \Phi(\mathcal{Z})+\lambda\Psi(\mathbf{E}) \nonumber\\
  &&\text{s.~t.} \quad \mathbf{X}_v=\mathbf{X}_v\mathbf{Z}_v+\mathbf{E}_v, v=1,\cdots,V,\\
  &&\quad\quad \mathcal{Z}=\Omega(\mathbf{Z}_1,\mathbf{Z}_2,\cdots,\mathbf{Z}_V),\mathbf{E} = [\mathbf{E}_1;\mathbf{E}_2;\cdots;\mathbf{E}_V],\nonumber
\end{eqnarray}
where the term $\Omega(\cdot)$ merges all self-representation matrices $\mathbf{Z}_v(v=1,\cdots, V)$ to a third-order tensor $\mathcal{Z}\in  \mathbb{R}^{N\times N\times V}$, $\Phi$ depicts a tensor rank term to explore the relationship of observed samples,  $\Psi$ is another term in accordance with corrupted noise type,
and $\lambda$ is a trade-off parameter to balance the effect of the $\Phi$ and  $\Psi$  regulations. Usually, sparse noise or outliers will be considered in subspace clustering. Hence, the corresponding convex surrogates $\ell_1$ or $\ell_{2,1}$ will be used to remove the sparse noise.
 
Tucker decomposition~\cite{kolda2009tensor} and t-SVD~\cite{wang2021low} are two commonly used tensor approximation methods for MSC.
For example, ~\cite{zhang2015low} considers Tucker decomposition and its corresponding tensor nuclear norm minimization to extract the low-rank information. Following it, ~\cite{chen2021low} considers simultaneously learning the low-rank Tucker self-representation tensor and the affinity matrix to preserve their correlations. In addition, there are many t-SVD based works with various tensor norms, e.g., standard t-SVD based tensor nuclear norm~\cite{xie2018unifying},  weighted tensor nuclear norm~\cite{gao2020tensor}, and weighted tensor Schatten $p$-norm~\cite{xia2021multiview}.  To further enhance the clustering performance, $\ell_{1,1,2}$ norm term can be added on the self-representation tensor to capture the local correlation in the view dimension~\cite{ yin2018multiview}.

\section{PROPOSED METHOD}

The self-representation tensor obtained from the current MSC works is mainly based on low-rank assumptions, which can explore the global relationship among views. However, the existing works are mainly based on Tucker or t-SVD, which can not fully explore the inter/intra-view information within the self-representation tensor.
In this work, we apply low-rank MERA decomposition for MSC, which can adaptively attain a well-explored structure of the self-representation tensor.


\subsection{Low-Rank MERA approximation}
\label{low_rank_MERA}
Low-rank tensor networks have been shown to be more effective in approximating high-order data~\cite{long2021bayesian, wang2017efficient, bengua2017efficient}, hence the self-representation tensor $\mathcal{Z}\in \mathbb{R}^{N\times N\times V}$ is rearranged into a 5th-order tensor
$\mathcal{Y}\in\mathbb{R}^{A\times Q\times A\times Q \times V}$ with $N=AQ$ for low-rank MERA approximation, while ensuring that $A$ and $Q$ are probably close to equilibrium.
In this part, we mainly give solutions of low-rank MERA factors for a 5th-order tensor in detail.

Given $\mathcal{Y}\in \mathbb{R}^{I_1\times\cdots\times I_5}$, its low-rank MERA approximation is to find MERA factors with predefined MERA ranks $R_1=R_2=R$, which can  be formulated as
\begin{equation}\label{update:MERA}
\min_{\mathcal{U},\mathcal{W},\mathbf{B}} \frac{1}{2}\| \mathcal{Y}-f(\mathbf{B},\mathcal{W},\mathcal{U})\|_{\operatorname{F}}^2.\end{equation}

As shown in Fig. \ref{fig:MERA}, there are four MERA factors including one disentangler $\mathcal{U}_1$,  two isometries $\mathcal{W}_1$, $\mathcal{W}_2$, and one top core $\mathbf{B}$. Each variable can be  alternately updated while the others are fixed. \\
\textbf{Update} $\mathcal{U}_1$:
With fixed $\mathcal{W}_1$,$\mathcal{W}_2$, and $\mathbf{B}$, the subproblem of $\mathcal{U}_1$ can be rewritten into a matrix form as:
\begin{equation}\label{update:u}
\min_{\mathbf{U}_1} \frac{1}{2}\| \mathbf{Y}_{(2, 3)}-\mathbf{U}_1\mathbf{M}_u\|_{\operatorname{F}}^2, \text{ s. t. } \mathbf{U}_1^{\operatorname{T}}\mathbf{U}_1=\mathbf{I}_{I_2 I_3},
\end{equation}
where $\mathbf{Y}_{(2, 3)}$ and $\mathbf{M}_u$ are the  mode-\{2, 3\} unfolding matrices of $\mathcal{Y}$ and $\mathcal{M}_u$, respectively. $\mathcal{M}_u$ is the tensor contraction result of other variables, i.e., $\mathcal{M}_u=\mathbf{B}\times_{\{R_1\}}\mathcal{W}_{1}\times_{\{R_2\}}\mathcal{W}_2\in\mathbb{R}^{I_1\times I_2 \times I_3 \times I_4 \times I_5}$; $\mathbf{U}_1$ is the mode-\{$1, 2$\} unfolding matrix of $\mathcal{U}_1\in\mathbb{R}^{I_2\times I_3\times I_2\times I_3}$.

The optimization problem (\ref{update:u}) is equivalent to  the following  well-known orthogonal Procrustes problem:
\begin{equation}\label{modelU}
    \max_{\mathbf{U}_1} \quad \operatorname{trace}((\mathbf{U}_1)^{\operatorname{T}}\mathbf{Y}_{(2,3)}(\mathbf{M}_u)^{\operatorname{T}}),
\end{equation}
where $\operatorname{trace}(\mathbf{A})$ calculates the sum of the diagonal elements of $\mathbf{A}$, which can be achieved by the Matlab command ``trace". 
In this way, the updating of $\mathcal{U}_1$ can be obtained by 
\begin{equation}\label{solution:U}
\mathcal{U}_1=\operatorname{fold}_{1, 2}(\mathbf{S}_u\mathbf{D}_u^{\operatorname{T}}),
\end{equation}
where $\mathbf{S}_u$ and $\mathbf{D}_u$ are the left and right singular vector matrices of the matrix $\mathbf{Y}_{(2, 3)}(\mathbf{M}_u)^{\operatorname{T}}$.\\
\textbf{Update} $\mathcal{W}_1$, $\mathcal{W}_2$:
Following the way of solving $\mathcal{U}_1$, the updating of $\mathcal{W}_p,~p = 1, 2$ can be obtained by
\begin{equation}\label{solution:W}
\mathcal{W}_p=\argmin_{\mathcal{W}_p:\mathbf{W}_p^{\operatorname{T}}\mathbf{W}_p=\mathbf{I}_{R_p}}\frac{1}{2}\| \mathcal{Y}-f(\mathbf{B},\mathcal{W},\mathcal{U})\|_{\text{F}}^2.
\end{equation}
\textbf{Update} $\mathbf{B}$: 
The subproblem of $\mathbf{B}$ can be rewritten as:
\begin{equation}
\min_{\mathbf{B}} \quad
\frac{1}{2}\| \mathcal{Y}-\mathcal{C}\times_{\{R_1, R_2\}}\mathbf{B}\|_{\operatorname{F}}^2,
\end{equation}
where $\mathcal{C}$ is a 7-th order layer in Definition 3. According to the orthogonal/semi-orthogonal property of disentangler $\mathcal{U}_1$ and isometries $\mathcal{W}_p, p = 1, 2$, the updating of $\mathbf{B}$ can be obtained by 
\begin{equation}\label{sol:B}
\mathbf{B}=\mathcal{Y}\times_{\{I_1,I_2,I_3,I_4,I_5\}}\mathcal{C}.
\end{equation}
Repeatedly update equations (\ref{solution:U}), (\ref{solution:W}),  and (\ref{sol:B})  for several iterations to achieve low-rank MERA approximation, which is summarized in Algorithm \ref{alg:algorithm1}.
\begin{algorithm}[H]
	\caption{Low-rank MERA approximation}	
           	\begin{algorithmic}[1]
			\STATE \textbf{Input:} 5-th order  data $\mathcal{Y}$, MERA ranks $R_1=R_2=R$
               \STATE \textbf{Initialize:} $\mathbf{B}$, $\mathcal{W}$, 
 $\mathcal{U}$,  Maximum iterations $S=10$
			\FOR{$s=1:S$ }
            \STATE Update $\mathcal{U}^{s}$ via equation (\ref{solution:U})
            \STATE Update $\mathcal{W}^{s}$ via equation (\ref{solution:W})
            \STATE Update $\mathbf{B}^{s}$ via equation (\ref{sol:B})
            \ENDFOR	
            \STATE $\hat{\mathcal{Y}}$=$f(\mathbf{B}^{S}$, $\mathcal{W}^{S}$, 
 $\mathcal{U}^{S}$)
			\STATE \textbf{Output:} Low-rank MERA approximation tensor $\hat{\mathcal{Y}}$
	\end{algorithmic}	
\label{alg:algorithm1}
\end{algorithm}

\subsection{Low-Rank MERA for MSC}
As shown in Fig. \ref{fig:framework1}, applying low-rank MERA decomposition on learned self-representation tensor $\mathcal{Z}$ for MSC can be formulated as the following optimization problem:
\begin{equation}\label{MERA model}
\begin{aligned}
&\min_{\{\mathbf{E}_v,\mathbf{Z}_v\}_{v=1}^{V}} \quad\sum_{v=1}^{V} \lambda\|\mathbf{E}_v\|_{2,1}\\
&\quad\text{s.~t.} \quad \mathbf{X}_v=\mathbf{X}_v\mathbf{Z}_v+\mathbf{E}_v, v=1,\cdots,V,\\
&\quad\qquad\hat{\mathcal{Z}}=f(\mathbf{B},\mathcal{W},\mathcal{U})
\end{aligned}
\end{equation}
where $\mathcal{Z}=\Omega\left(\mathbf{Z}_{1}, \mathbf{Z}_{2}, \cdots, \mathbf{Z}_{V}\right)\in\mathbb{R}^{N\times N\times V}$, $\hat{\mathcal{Z}}=\operatorname{reshape}(\mathcal{Z},[A,Q,A,Q,V])$. We use operator $\mathfrak{R}$
to represent these two steps. And  $\mathfrak{R}^{-1}$ denotes its inverse operator. $\hat{\mathcal{Z}}=f(\mathbf{B},\mathcal{W},\mathcal{U})$ means applying low-rank MERA approximation to explore both inter-view and intra-view information.


\subsection{Solutions}
To make the above optimization problem separable,  an auxiliary variable $\mathcal{Y}$ is introduced as follows:
\begin{equation}\label{MERA model1}
\begin{aligned}
&\min_{\{\mathbf{E}_v,\mathbf{Z}_v\}_{v=1}^{V}} \quad\sum_{v=1}^{V} \lambda\|\mathbf{E}_v\|_{2,1}\\
&\quad\text{s.~t.} \quad \mathbf{X}_v=\mathbf{X}_v\mathbf{Z}_v+\mathbf{E}_v, v=1,\cdots,V,\\
&\quad\qquad\hat{\mathcal{Z}}=\mathcal{Y}, \mathcal{Y}=f(\mathbf{B},\mathcal{W},\mathcal{U}).
\end{aligned}
\end{equation}
The Lagrangian function can be formulated as follows:
\begin{equation}
\begin{aligned}
&\operatorname{L}\left(\{\mathbf{Z}_{v},\mathbf{E}_v,\Lambda_v\}_{v=1}^{V},\mathcal{Y},\Gamma\right) \\
&=\sum_{v=1}^{V} (\lambda\|\mathbf{E}_v\|_{2,1} +\langle\Lambda_{v}, \mathbf{X}_{v}-\mathbf{X}_{v}\mathbf{Z}_{v}-\mathbf{E}_{v}\rangle\\
&+\frac{\mu_2}{2}\|\mathbf{X}_{v}-\mathbf{X}_{v} \mathbf{Z}_{v}-\mathbf{E}_{v}\|_{\operatorname{F}}^{2})
+\frac{\mu_1}{2}\|\hat{\mathcal{Z}}-\mathcal{Y}\|_{\operatorname{F}}^{2}+\langle\Gamma ,\hat{\mathcal{Z}}-\mathcal{Y}\rangle,
\end{aligned}
\label{eq:alm}
\end{equation}
under the constraint $\mathcal{Y}=f(\mathbf{B},\mathcal{W},\mathcal{U})$, where $\{\Lambda_v\}_{v=1}^{V}$ and $\Gamma$ are Lagrangian multipliers and $\mu_1,\mu_2$ are penalty factors. According to the ADMM framework, problem (\ref{eq:alm}) can be divided into several subproblems where each variable can be alternately updated with other variables fixed.\\
\textbf{Update $\{\mathbf{Z}_v\}_{v=1}^{V}$}:
With other variables fixed, the solution of $\mathbf{Z}_v$ can be obtained by setting the derivative of the objective function in (\ref{eq:alm}) with respect to $\mathbf{Z}_{v}$ to zero, as follows:
\begin{equation}\label{sol:Z1}
\begin{aligned}
&-\mathbf{X}_v^{\operatorname{T}}\Lambda_v-
\mu_2\mathbf{X}_v^{\operatorname{T}}(\mathbf{X}_{v}-\mathbf{X}_{v} \mathbf{Z}_{v}-\mathbf{E}_{v})\\
&+\mu_1(\mathfrak{R}_v^{-1}(\hat{\mathcal{Z}}-\mathcal{Y})+\mathfrak{R}_v^{-1}(\Gamma))=\mathbf{0},
\end{aligned}
\end{equation}
where $\mathfrak{R}_v^{-1}$ is the inverse operator along the $v$-th view, i.e., $\mathfrak{R}_v^{-1}(\hat{\mathcal{Z}})=\mathbf{Z}_{v}$.
The solution of $\mathbf{Z}_v$ can be rewritten as:
\begin{equation}\label{sol:Z}
\begin{aligned}
\mathbf{Z}_{v}=&(\mu_1\mathbf{I}+\mu_2 \mathbf{X}_{v}^{\operatorname{T}} \mathbf{X}_{v})^{-1}(\mathbf{X}_{v}^{\operatorname{T}} \Lambda_{v}+\mu_2 \mathbf{X}_{v}^{\operatorname{T}} \mathbf{X}_{v}-\\
&\mu_2 \mathbf{X}_{v}^{\operatorname{T}} \mathbf{E}_{v}+\mathfrak{R}_v^{-1}(\mu_1\mathcal{Y}-\Gamma)).
\end{aligned}
\end{equation}\\
\textbf{Update $\{\mathbf{E}_v\}_{v=1}^{V}$}:
The solution of $\mathbf{E}_v$ can be obtained by optimizing
\begin{equation}
   \frac{\lambda}{\mu_2} \|\mathbf{E}_v\|_{2,1}+\frac{1}{2}\|\mathbf{E}_v-(\mathbf{X}_{v}-\mathbf{X}_{v} \mathbf{Z}_{v}+(1 / \mu_2) \Lambda_{v})\|_{\operatorname{F}}^2.
\end{equation}
Letting $\mathbf{D}=\mathbf{X}_{v}-\mathbf{X}_{v} \mathbf{Z}_{v}+(1 / \mu_2) \Lambda_{v}$,  and according to \cite{tang2018learning}, we have
\begin{equation}\label{sol:E}
\mathbf{E}_v(:,n)=\left\{\begin{array}{l}
\frac{\left\|\mathbf{D}(:,n)\right\|_{2}-\frac{\lambda}{\mu_2}}{\left\|\mathbf{D}(:,n)\right\|_{2}} \mathbf{D}(:,n), \quad\left\|\mathbf{D}(:,n)\right\|_{2}>\frac{\lambda}{\mu_2} \\
\mathbf{0}, ~~~~~~~~\qquad\qquad\qquad \text { otherwise,}
\end{array}\right.
\end{equation}
where $n=1\cdots,N$, $N$ is the number of samples.\\
\textbf{Update $\mathcal{Y}$}:
With respect to $\mathcal{Y}$, the problem (\ref{eq:alm}) can be transferred into the following formulation:
\begin{equation}\label{sol:Y}
\begin{aligned}
&\min _{\mathcal{Y}}\quad \frac{\mu_1}{2}\left\|\mathcal{Y}-\left(\hat{\mathcal{Z}}+\frac{1}{\mu_1} \Gamma\right)\right\|_{\operatorname{F}}^{2} \\
&~\text {s. t. }\quad \mathcal{Y}=f(\mathbf{B},\mathcal{W},\mathcal{U}).\\
\end{aligned}
\end{equation}
It can be solved by Algorithm \ref{alg:algorithm1}, where the input self-representation tensor is $\hat{\mathcal{Z}}+\frac{1}{\mu_1} \Gamma$.\\
\textbf{Update Lagrangian multipliers}:
\begin{equation}
\Gamma=\Gamma+\mu_1(\hat{\mathcal{Z}}-\mathcal{Y}).
\label{eq:otnmsc5}
\end{equation}
\begin{equation}
\Lambda_{v}=\Lambda_{v}+\mu_2\left(\mathbf{X}_{v}-\mathbf{X}_{v} \mathbf{Z}_{v}-\mathbf{E}_{v}\right), v = 1,\cdots,V.
\label{eq:otnmsc4}
\end{equation}

After finishing updating the variables, the affinity matrix of multi-view data can be obtained by $\mathbf{S}=\frac{1}{V} \sum_{v=1}^{V}\left|\mathbf{Z}_{v}\right|+\left|\mathbf{Z}_{v}^{\operatorname{T}}\right|$, which will be used in spectral clustering algorithm\cite{von2007tutorial} for the final clustering result. The optimization processing of MERA-MSC is summarized in algorithm \ref{alg:algorithm2}.  
When  $ \max_{v,v=1,\cdots,V} \left\| \mathbf{X}_v-\mathbf{X}_v\mathbf{Z}^t_v-\mathbf{E}^t_v\right\|_{\infty} \leq \varepsilon$ and $ \max_{v,v=1,\cdots,V} \left\| \mathbf{Z}^t_v-\mathbf{Y}^t_v\right\|_{\infty} \leq \varepsilon $, $\varepsilon=10^{-6}$, our algorithm stops. 

The computational complexity of the MERA-MSC algorithm for one iteration is summarized in Table \ref{tab:data}. Overall, the main computational complexity is $\operatorname{O}(TN^3(V+1))$ where $T$ is the number of iterations in Algorithm \ref{alg:algorithm2}.
\begin{algorithm}[H]
	\caption{MERA-MSC}	
           	\begin{algorithmic}[1]
			\STATE \textbf{Input:} Multi-view data $\{\mathbf{X}_{v}\}_{v=1}^V, \lambda$
			\STATE \textbf{Initialize:} $\mathcal{Y} = \Gamma = 0; \mathbf{Z}_{v} = \mathbf{E}_v = \Lambda_v = \mathbf{0}, v = 1, \cdots, V; \mu_1 = 10^{-4}; \mu_2 = 5^{-4};
                 \varepsilon = 10^{-6}; \eta = 2; \mu_1^{\max}=\mu_2^{\max } = 10^{10}; t = 1, T = 50$
			\WHILE{ $ t\leq T$ }
            \FOR{$v=1:V$}
            \STATE Update $\mathbf{Z}_{v}^{t}$ via equation (\ref{sol:Z})
            \STATE Update $\mathbf{E}_v^{t}$ via equation (\ref{sol:E})
            \STATE Update $\Lambda_v^{t}$ via equation (\ref{eq:otnmsc4})
            \ENDFOR
            \STATE Update $\mathcal{Y}^t$ via equation (\ref{sol:Y})
            \STATE Update $\Gamma^t$ by equation (\ref{eq:otnmsc5})
            \STATE  $\mu_1 = \min \left(\eta \mu_1, \mu_1^{\max }\right)$, $\mu_2 = \min \left(\eta \mu_2, \mu_2^{\max }\right)$;
            \STATE Check convergence conditions   
            \ENDWHILE	
			\STATE Calculate affinity matrix: $\mathbf{S}=\frac{1}{V} \sum_{v=1}^{V}\left|\mathbf{Z}_{v}\right|+\left|\mathbf{Z}_{v}^{\operatorname{T}}\right|$
			\STATE Apply spectral clustering algorithm using $\mathbf{S}$
			\STATE \textbf{Output:} Clustering result
	\end{algorithmic}	
\label{alg:algorithm2}
\end{algorithm}

\begin{table}[htbp]
\caption{The computational complexity for one iteration, where $D$, $N$, and $V$ are the dimension of features, the number of samples, and views, respectively.}
\small
\centering
    \begin{tabular}{{cccccc}}
		\toprule
	Variables&	 $\mathbf{Z}_v$  	& $\mathbf{E}_v$ 	& $\mathcal{Y}$  	\\
		\midrule
	Complexity	&$\operatorname{O}(N^3)$	& $\operatorname{O}(DN^2)$ 	&  $\operatorname{O}(N^3V)$ 	\\
          \midrule
      Main Complexity  & \multicolumn{2}{r}{$\operatorname{O}(N^3(1+V))$}  	\\
		\bottomrule	
	\end{tabular}
\label{tab:data}
\end{table}


\vspace{-0.5cm}

\section{EXPERIMENTS}
\subsection{Experimental Settings}
\subsubsection{Multi-view Datasets Description}
Five well-known multi-view datasets are chosen to evaluate the effectiveness of MERA-MSC:
   \textbf{Yale}\footnote{http://vision.ucsd.edu/content/yale-face-database} contains 165 samples of 15 clusters, with 11 samples per subject. We choose 3304-dimension(D) LBP, 6750-D Gabor, and 4096-D intensity as 3 views. 
    \textbf{MSRC-v5}\cite{winn2005locus} consists of 210 image samples collected from 7 clusters with 5 views, including 254-D CENT, 24-D CMT, 512-D GIST, 576-D HOG, 256-D LBP. 
    \textbf{Extended YaleB}\footnote{http://vision.ucsd.edu/leekc/ExtYaleDatabase/ExtYaleB.html} collects 2432 face image samples from 38 individuals, where each one has 64 images. Following~\cite{jia2021multi}, we choose the first 10 classes with 3 views, including 1024-D intensity, 1239-D LBP, and 256-D Gabor. Since data points in one subspace are very close to other subspaces, clustering on this dataset can be a challenge~\cite{elhamifar2013sparse}. 
    \textbf{Notting-Hill}~\cite{zhang2009character} contains 4660 face samples with 5 classes. We choose intensity 2000-D, 3304-D LBP, and 6750-D Gabor as 3 views. Due to the large number of samples, we will downsample it as 2330 samples with 5 classes.
    \textbf{BDGP}~\cite{cai2012joint} contains 2500 samples from 5 classes with 4 views, including 1000-D lateral, 500-D dorsal, 250-D ventral, and 79-D texture.
    
 The  statistical information of the above datasets is summarized in Table \ref{tab:data}. Note that each self-representation tensor of size $N \times N \times V$ is rearranged into a 5th-order tensor for low-rank MERA approximation. The last column of Table \ref{tab:data} displays the sizes of the 5th-order self-representation tensors corresponding to different datasets.
More information about this rearrangement can be found in Subsection \ref{low_rank_MERA}.
\begin{table}[htbp]
\small
\centering
\caption{ Statistical information of different multi-view datasets.}
 \resizebox{\linewidth}{!}{
	\begin{tabular}{{cccccc}}
	\toprule \toprule \noalign{\smallskip}
		Datasets  	&Samples 	&Views  	&Clusters  	&5-D ($I_{1},\cdots,I_{5}$)\\
		\midrule
		Yale		&165  	&3   	&15 		&(11,15,15,11,3)\\
		MSRC-v5 &210  &5 &7   &(15,14,15,14,5)\\
       Extended YaleB    &640    &3      &10        &(32,20,20,32,3)\\
  		Notting-Hill    &2330    &3      &5        &(233,10,233,10,3)\\
       BDGP    &2000    &4      &5         &(50,50,50,50,4)\\
			\midrule
  \bottomrule	
	\end{tabular}}
\label{tab:data}
\end{table}
\subsubsection{Compared Clustering Algorithms}
In order to compare clustering performance, we select nine state-of-the-art methods including one single-view method: LRR$_{best}$~\cite{liu2012robust}, seven self-representation based multi-view methods: multi-view low-rank representation (MLRR) [2021, TCYB]~\cite{chen2021multiview}, latent multi-view subspace clustering(L-MSC) [2017, CVPR]~\cite{zhang2017latent}, exclusivity-consistency regularized multi-view subspace clustering (ECMSC) [2017, CVPR]~\cite{wang2017exclusivity}, low-rank tensor constrained multi-view subspace clustering (LTMSC) [2015, ICCV]~\cite{zhang2015low}, t-SVD based multi-view subspace clustering (t-SVD-MSC) [2018, IJCV]~\cite{xie2018unifying}, weighted tensor Schatten $p$-norm based multi-view subspace clustering (WTSNM) [2021, TCYB]~\cite{xia2021multiview}, multi-view subspace clustering tailored tensor low-rank representation (MVSC-TLRR) [2021, TCSVT]~\cite{jia2021multi}, and one updated tensor based multi-view clustering method: tensorized bipartite graph learning for multi-view clustering (TBGL-MVC) [2022, TPAMI]~\cite{xia2022tensorized}. All tests are accomplished on a desktop computer with 2.4 GHz Quad-Core Intel Core i5 Processor and 16 GB 2133 MHz LPDDR3 Memory.
\subsubsection{Evaluation Metrics}
Six standard evaluation metrics are considered to evaluate the performance in our experiments, including F-score, Precision, Recall, normalized mutual information (NMI), adjusted rand index (ARI), and accuracy (ACC).  The larger values of these metrics indicate better clustering performance.  The detailed information can refer to~\cite{schutze2008introduction}.

\subsubsection{Parameter settings}
\begin{figure}[htbp]
 \centering
 \begin{subfigure}[b]{0.44\textwidth}
 \includegraphics[width=\textwidth]{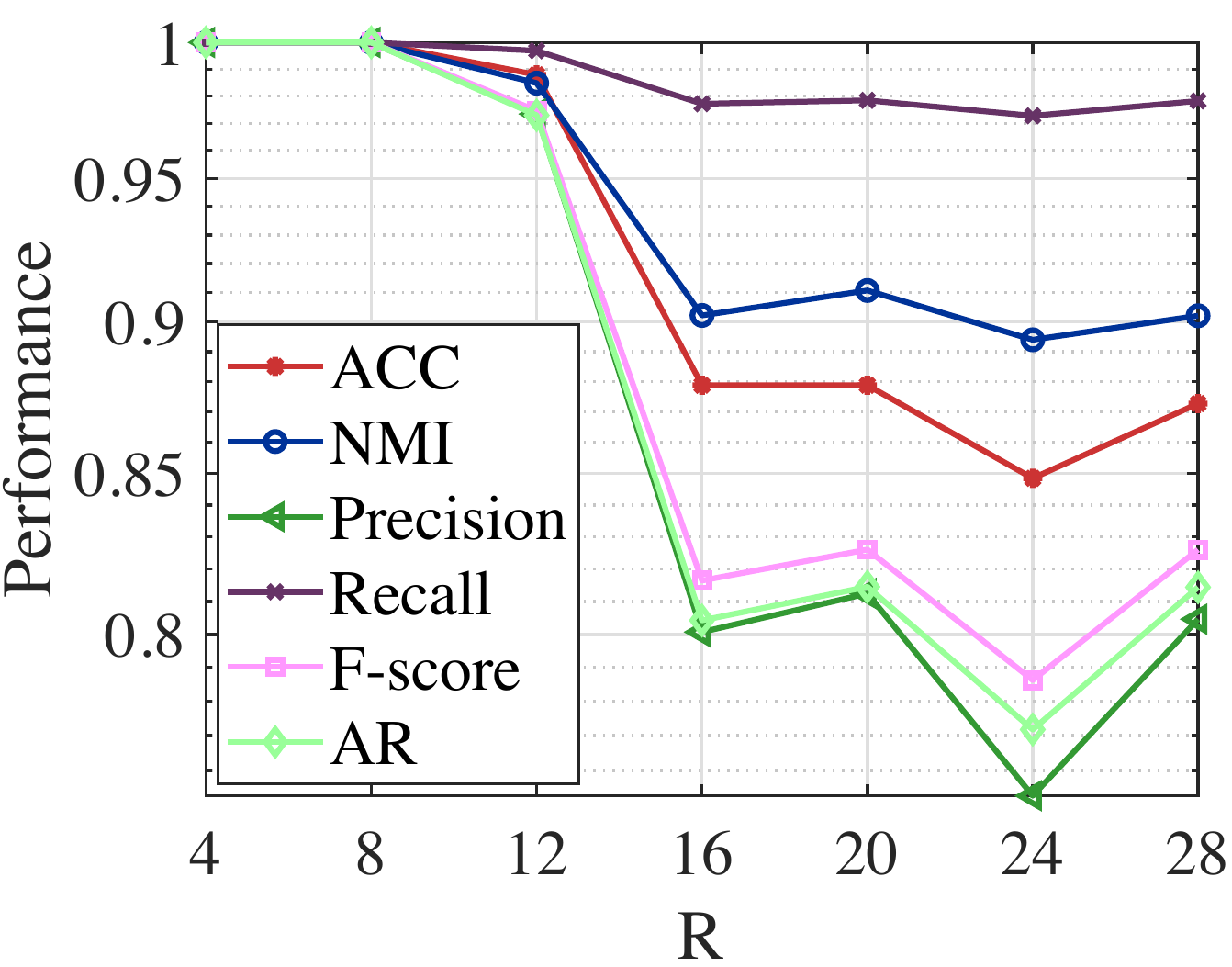}
    \caption{ $\lambda=0.001$}
    \label{fig:first1}
\end{subfigure}
\begin{subfigure}[b]{0.435\textwidth}
   \includegraphics[width=\textwidth]{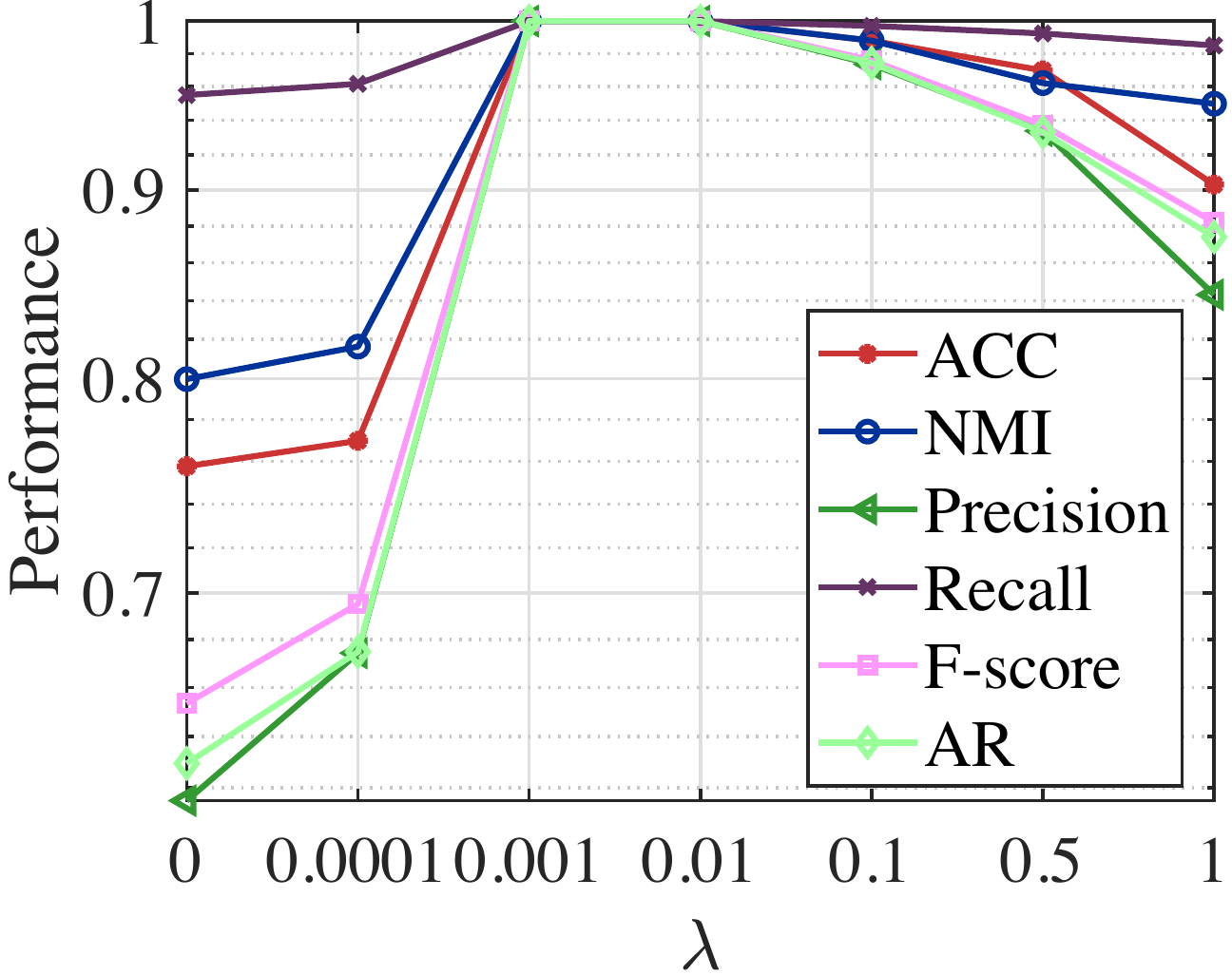}
    \caption{$R=6$}
    \label{fig:second1}
\end{subfigure}
\caption{The change of clustering performance as parameters $\lambda$ and $R$ vary. (a) $R$ changed with $\lambda=0.001$; (b) $\lambda$ changed with $R=6$. }
\label{fig:Rrho}
 \end{figure}

Our MERA-MSC has two free parameters, one for MERA decomposition, namely MERA rank $R$, and the other is the balancing parameter $\lambda$. We fix one and tune the other via brute force search.
For example, we first fix $\lambda = 0.01$ and tune parameters $R$ on the Yale dataset,
where $R$ varies from \{4, 8, 12, 16, 20, 24, 28\}. The clustering results are reported in Fig. \ref{fig:Rrho}(\subref{fig:first1}), where the ACC almost approaches 1 and performs stably when $R$ ranges in $[4,8]$.
In addition, we fix $R = 6$ to choose the parameter $\lambda$ from \{0, 0.0001, 0.01, 0.1, 0.5, 1\}.
From Fig. \ref{fig:Rrho}(\subref{fig:second1}), we can observe the best clustering performance in terms of six metrics is achieved in a wide range of $\lambda$, i.e., $\lambda\in [0.001,0.1]$.

\subsection{Clustering Performance Analysis}
Table \ref{tab:yale} shows the clustering performance of different methods on five multi-view datasets in terms of F-score, Precision, Recall, AR, NMI, and ACC, where the best and the second best results are highlighted  in \textbf{bold} and  \underline{underlined}, respectively.
For MSRC datasets, several multi-view methods, including WTSNM, MVSC-TLRR, TBGL-MVC, and MERA-MSC, all reached the best performance with all metrics achieved 1. Meanwhile, our MERA-MSC also gives the best performance on Yale datasets and outperforms all compared baselines, especially on the extended YaleB dataset. Specifically, our MERA-MSC provides 2.7\%  Precision improvement on the Yale dataset; 6.9\% NMI improvement on the extended YaleB dataset, 6.6\% AR improvement on the Notting Hill dataset, and 0.4\% F-score on the BDGP dataset.

Moreover, tensor-based MSC clustering methods, including t-SVD-MSC, WTSNM, MVSC-TLRR and MERA-MSC, generally perform better than matrix-based ones. 
  It verifies the capability of low-rank tensor decomposition to explore high-order correlations from different views. 
 However, considering the extended YaleB dataset
 , which is hard to cluster due to the effects of high illumination variation; only MVSC-TLRR and the proposed MERA-MSC perform well.
Compared to other tensor-based MSC methods, these two methods investigate both inter- and intra-view information throughout the self-representation tensor learning process. The introduction of intra-view information exploration can further enhance the clustering performance. 
In addition, compared with MVSC-TLRR, our method performs better. This is mainly because low-rank MERA decomposition can adaptively learn both inter-/intra-view from the self-representation tensor and provides a powerful and flexible representation. But for MVSC-TLRR, the contributions from inter-view and intra-view are required to be balanced by weight selection  and thus reduce its flexibility.  
 
 
\begin{table*}[!t]
\caption{Clustering results, e.g., mean value (Standard Deviation) on Yale, MSRC-v5, Extended YaleB, Notting Hill, and BDGP datasets.}
\small
\begin{center}\resizebox{\linewidth}{!}{
\begin{tabular}{{c|c|c|c|c|c|c|c c}}
\toprule 
  Datasets 	&  Methods 	&  F-score 	&  Precision 	&  Recall 	&    NMI 	&     ARI 	&    ACC 	\\ 
\midrule 
 \multirow{12}{*}{\shortstack{MSRC-v5\\($R$=2,$\lambda$=0.001)}} 
 & LRR-best	 &0.5469(0.0025)	 &0.5469(0.0027)	 &0.5469(0.0027)	 &0.5658(0.0049)	 &0.4734(0.0029)	 &0.6862(0.0015)	  \\ 	
&   MLRR	 &0.6674(0.0035)	 &0.6604(0.0031)	 &0.6745(0.0031)	 &0.6828(0.0028)	 &0.6131(0.0040)	 &0.7800(0.0038)	  \\ 		
&  L-MSC	 &0.6351(0.0074)	 &0.6197(0.0087)	 &0.6512(0.0087)	 &0.6597(0.0086)	 &0.5745(0.0088)	 &0.7714(0.0045)	  \\ 	
&  ECMSC	 &0.6924(0.0000)	 &0.6804(0.0000)	 &0.7048(0.0000)	 &0.7185(0.0000)	 &0.6418(0.0000)	 &0.7857(0.0000)	  \\ 	
&  LTMSC	 &0.7148(0.0035)	 &0.7023(0.0036)	 &0.7276(0.0036)	 &0.7444(0.0037)	 &0.6679(0.0041)	 &0.8338(0.0027)	  \\ 	
& tSVD-MSC	 &\underline{0.9806(0.0000)}	 &\underline{0.9803(0.0000)}	 &\underline{0.9810(0.0000)}	 &\underline{0.9785(0.0000)}	 &\underline{0.9775(0.0000)}	 &\underline{0.9905(0.0000)}	 \\ 	
 & WTSNM	&\textbf{1.0000(0.0000)}	 &\textbf{1.0000(0.0000)}	 &\textbf{1.0000(0.0000)}	 &\textbf{1.0000(0.0000)}	 &\textbf{1.0000(0.0000)}	 &\textbf{1.0000(0.0000)}	 \\ 	
 &MVSC-TLRR	 &\textbf{1.0000(0.0000)}	 &\textbf{1.0000(0.0000)}	 &\textbf{1.0000(0.0000)}	 &\textbf{1.0000(0.0000)}	 &\textbf{1.0000(0.0000)}	 &\textbf{1.0000(0.0000)}	 \\ 	
 &TBGL-MVC	&\textbf{1.0000(0.0000)}	 &\textbf{1.0000(0.0000)}	 &\textbf{1.0000(0.0000)}	 &\textbf{1.0000(0.0000)}	 &\textbf{1.0000(0.0000)}	 &\textbf{1.0000(0.0000)}	 \\  	
 &MERA-MSC	 &\textbf{1.0000(0.0000)}	 &\textbf{1.0000(0.0000)}	 &\textbf{1.0000(0.0000)}	 &\textbf{1.0000(0.0000)}	 &\textbf{1.0000(0.0000)}	 &\textbf{1.0000(0.0000)}	 \\ 			
\bottomrule
 \multirow{12}{*}{\shortstack{Yale\\($R$=6,$\lambda$=0.001)}}  
& LRR-best	 &0.5483(0.0212)	 &0.5483(0.0202)	 &0.5483(0.0202)	 &0.7111(0.0146)	 &0.5180(0.0226)	 &0.7061(0.0146)	  \\ 	
&   MLRR	 &0.4884(0.0154)	 &0.4725(0.0145)	 &0.5055(0.0145)	 &0.6569(0.0112)	 &0.4540(0.0163)	 &0.6327(0.0192)	  \\ 	
&  L-MSC	 &0.5144(0.0096)	 &0.4689(0.0129)	 &0.5699(0.0129)	 &0.7131(0.0058)	 &0.4796(0.0107)	 &0.6709(0.0076)	  \\ 	
&  ECMSC	 &0.4843(0.0000)	 &0.4428(0.0000)	 &0.5345(0.0000)	 &0.7029(0.0000)	 &0.4475(0.0000)	 &0.6545(0.0000)	  \\ 	
&  LTMSC	 &0.6005(0.0114)	 &0.5785(0.0122)	 &0.6242(0.0122)	 &0.7500(0.0080)	 &0.5735(0.0122)	 &0.7297(0.0042)	  \\ 	
& tSVD-MSC	 &0.9155(0.0657)	 &0.9037(0.0753)	 &0.9280(0.0753)	 &0.9536(0.0364)	 &0.9099(0.0701)	 &0.9394(0.0525)	  \\ 	
&  WTSNM	 &0.9725(0.0313)	 &0.9628(0.0504)	 &0.9833(0.0504)	 &0.9875(0.0104)	 &0.9706(0.0335)	 &0.9800(0.0294)	  \\ 	
& MVSC-TLRR	 &\underline{0.9746(0.0000)}	 &\underline{0.9734(0.0000)}	 &\underline{0.9758(0.0000)}	 &\underline{0.9848(0.0000)}	 &\underline{0.9729(0.0000)}	 &\underline{0.9879(0.0000)}	 \\ 		
& TBGL-MVC	 &0.6475(0.0000)	 &0.5626(0.0000)	 &0.7624(0.0000)	 &0.8531(0.0000)	 &0.6208(0.0000)	 &0.8606(0.0000)	  \\ 	
 &MERA-MSC	 &\textbf{1.0000(0.0000)}	 &\textbf{1.0000(0.0000)}	 &\textbf{1.0000(0.0000)}	 &\textbf{1.0000(0.0000)}	 &\textbf{1.0000(0.0000)}	 &\textbf{1.0000(0.0000)}	 \\ 		 	
\bottomrule 
 \multirow{12}{*}{\shortstack{Extended YaleB\\($R$=10,$\lambda$=1)}} 
  & LRR-best	 &0.4750(0.0008)	 &0.4750(0.0007)	 &0.4750(0.0007)	 &0.6014(0.0013)	 &0.4163(0.0009)	 &0.5827(0.0005)	  \\ 	
&   MLRR	 &0.1789(0.0021)	 &0.1756(0.0023)	 &0.1822(0.0023)	 &0.1889(0.0035)	 &0.0872(0.0025)	 &0.2561(0.0058)	  \\ 
&  L-MSC	 &0.2791(0.0023)	 &0.2291(0.0028)	 &0.3571(0.0028)	 &0.4453(0.0035)	 &0.1807(0.0030)	 &0.4644(0.0010)	  \\ 	
&  ECMSC	 &0.3707(0.0000)	 &0.2841(0.0000)	 &0.5332(0.0000)	 &0.6282(0.0000)	 &0.2777(0.0000)	 &0.6312(0.0000)	  \\ 	
&  LTMSC	 &0.4154(0.0049)	 &0.4063(0.0055)	 &0.4250(0.0055)	 &0.5424(0.0040)	 &0.3499(0.0056)	 &0.5109(0.0051)	  \\ 	
& tSVD-MSC	 &0.4584(0.0050)	 &0.4231(0.0053)	 &0.5001(0.0053)	 &0.6113(0.0040)	 &0.3936(0.0057)	 &0.6005(0.0028)	  \\ 	
&  WTSNM	 &0.4426(0.0043)	 &0.4070(0.0047)	 &0.4851(0.0047)	 &0.5964(0.0033)	 &0.3757(0.0050)	 &0.5875(0.0028)	  \\ 	
& MVSC-TLRR	 &\underline{0.9139(0.0013)}	 &\underline{0.9125(0.0013)}	 &\underline{0.9153(0.0013)}	 &\underline{0.9123(0.0009)}	 &\underline{0.9045(0.0014)}	 &\underline{0.9558(0.0008)}	 \\ 	
& TBGL-MVC	 &0.2394(0.0000)	 &0.1560(0.0000)	 &0.5140(0.0000)	 &0.4984(0.0000)	 &0.1038(0.0000)	 &0.4641(0.0000)	  \\ 	
 &MERA-MSC	 &\textbf{0.9535(0.0000)}	 &\textbf{0.9497(0.0000)}	 &\textbf{0.9574(0.0000)}	 &\textbf{0.9653(0.0000)}	 &\textbf{0.9484(0.0000)}	 &\textbf{0.9766(0.0000)}	 \\ 		
\bottomrule 
 \multirow{10}{*}{\shortstack{Notting Hill\\($R$=15,$\lambda$=0.0001)}} 
& LRR-best	 &0.8047(0.0000)	 &0.8047(0.0000)	 &0.8047(0.0000)	 &0.7456(0.0000)	 &0.7520(0.0000)	 &0.8811(0.0000)	  \\ 	
&   MLRR	 &0.8394(0.0008)	 &0.8301(0.0006)	 &0.8490(0.0006)	 &0.8006(0.0008)	 &0.7936(0.0010)	 &0.8446(0.0007)	  \\ 	
&  L-MSC	 &0.7967(0.0000)	 &0.8033(0.0000)	 &0.7903(0.0000)	 &0.8250(0.0000)	 &0.7402(0.0000)	 &0.7777(0.0000)	  \\ 	
&  ECMSC	 &0.6684(0.0000)	 &0.5455(0.0000)	 &0.8627(0.0000)	 &0.6919(0.0000)	 &0.5465(0.0000)	 &0.7300(0.0000)	  \\ 		
&  LTMSC	 &0.8320(0.0003)	 &0.8386(0.0004)	 &0.8255(0.0004)	 &0.7949(0.0004)	 &0.7853(0.0004)	 &0.9036(0.0002)	  \\ 	
& tSVD-MSC	 &0.8901(0.0000)	 &0.9096(0.0000)	 &0.8715(0.0000)	 &0.8743(0.0000)	 &0.8601(0.0000)	 &0.9421(0.0000)	  \\ 	
&  WTSNM	 &\underline{0.9101(0.0000)}	 &\underline{0.9280(0.0000)}	 &\underline{0.8929(0.0000)}	 &\underline{0.8939(0.0000)}	 &\underline{0.8854(0.0000)}	 &\underline{0.9532(0.0000)}	 \\ 	
& MVSC-TLRR	 &0.8766(0.0000)	 &0.8929(0.0000)	 &0.8609(0.0000)	 &0.8555(0.0000)	 &0.8427(0.0000)	 &0.9343(0.0000)	  \\ 	
& TBGL-MVC	 &0.7586(0.0000)	 &0.7436(0.0000)	 &0.7742(0.0000)	 &0.8123(0.0000)	 &0.6890(0.0000)	 &0.8476(0.0000)	  \\ 	
 &MERA-MSC	 &\textbf{0.9619(0.0000)}	 &\textbf{0.9552(0.0000)}	 &\textbf{0.9687(0.0000)}	 &\textbf{0.9450(0.0000)}	 &\textbf{0.9511(0.0000)}	 &\textbf{0.9700(0.0000)}	 \\ 	
\bottomrule 
 \multirow{10}{*}{\shortstack{BDGP\\($R$=10,$\lambda$=0.0002)}} 
& LRR-best	 &0.7448(0.0000)	 &0.7448(0.0000)	 &0.7448(0.0000)	 &0.8020(0.0000)	 &0.6711(0.0000)	 &0.7004(0.0000)	  \\ 	
&   MLRR	 &0.6838(0.0000)	 &0.6113(0.0000)	 &0.7758(0.0000)	 &0.7006(0.0000)	 &0.5928(0.0000)	 &0.7064(0.0000)	  \\ 	
&  L-MSC	 & 0.7354(0.0000)	 & 0.6673(0.0000)	 & 0.8191(0.0000)	 & 0.7762(0.0000)	 & 0.6601(0.0000)	 & 0.6912(0.0000)	  \\ 	
&  ECMSC	 & 0.7562(0.0000)	 & 0.717(0.0000)	 & 0.800(0.0000)	 & 0.761(0.0000)	 & 0.691(0.0000)	 & 0.789(0.0000)	  \\ 	
&  LTMSC	 &0.4325(0.0000)	 &0.4281(0.0000)	 &0.4369(0.0000)	 &0.3889(0.0000)	 &0.2891(0.0000)	 &0.5272(0.0000)	  \\ 	
& tSVD-MSC	 &0.9889(0.0000)	 &0.9888(0.0000)	 &0.9890(0.0000)	 &0.9814(0.0000)	 &0.9861(0.0000)	 &0.9944(0.0000)	  \\ 			
&  WTSNM	 &\underline{ 0.9901(0.0000)}	 &\underline{ 0.9901(0.0000)}	 &\underline{ 0.9911(0.0000)}	 &\underline{ 0.9840(0.0000)}	 &\underline{ 0.9881(0.0000)}	 &\underline{ 0.9951(0.0000)}	 \\ 	 	
& MVSC-TLRR	 &0.9842(0.0000)	 &0.9841(0.0000)	 &0.9842(0.0000)	 &0.9719(0.0000)	 &0.9802(0.0000)	 &0.9920(0.0000)	  \\ 		
& TBGL-MVC	 &0.3303(0.0000)	 &0.2000(0.0000)	 &0.9487(0.0000)	 &0.0410(0.0000)	 &0.0007(0.0000)	 &0.2236(0.0000)	  \\ 		
 &MERA-MSC	 &\textbf{ 0.9913(0.000)}	 &\textbf{ 0.9912(0.0000)}	 &\textbf{ 0.9913(0.0000)}	 &\textbf{ 0.9860(0.0000)}	 &\textbf{ 0.9891(0.0000)}	 &\textbf{ 0.9956(0.0000)}	 \\ 	
\bottomrule 
\end{tabular}}
\end{center}
\label{tab:yale}
\vspace{-0.5cm}
\end{table*}

 \begin{figure}[ht!]
 \centering
\includegraphics[width=1\textwidth]{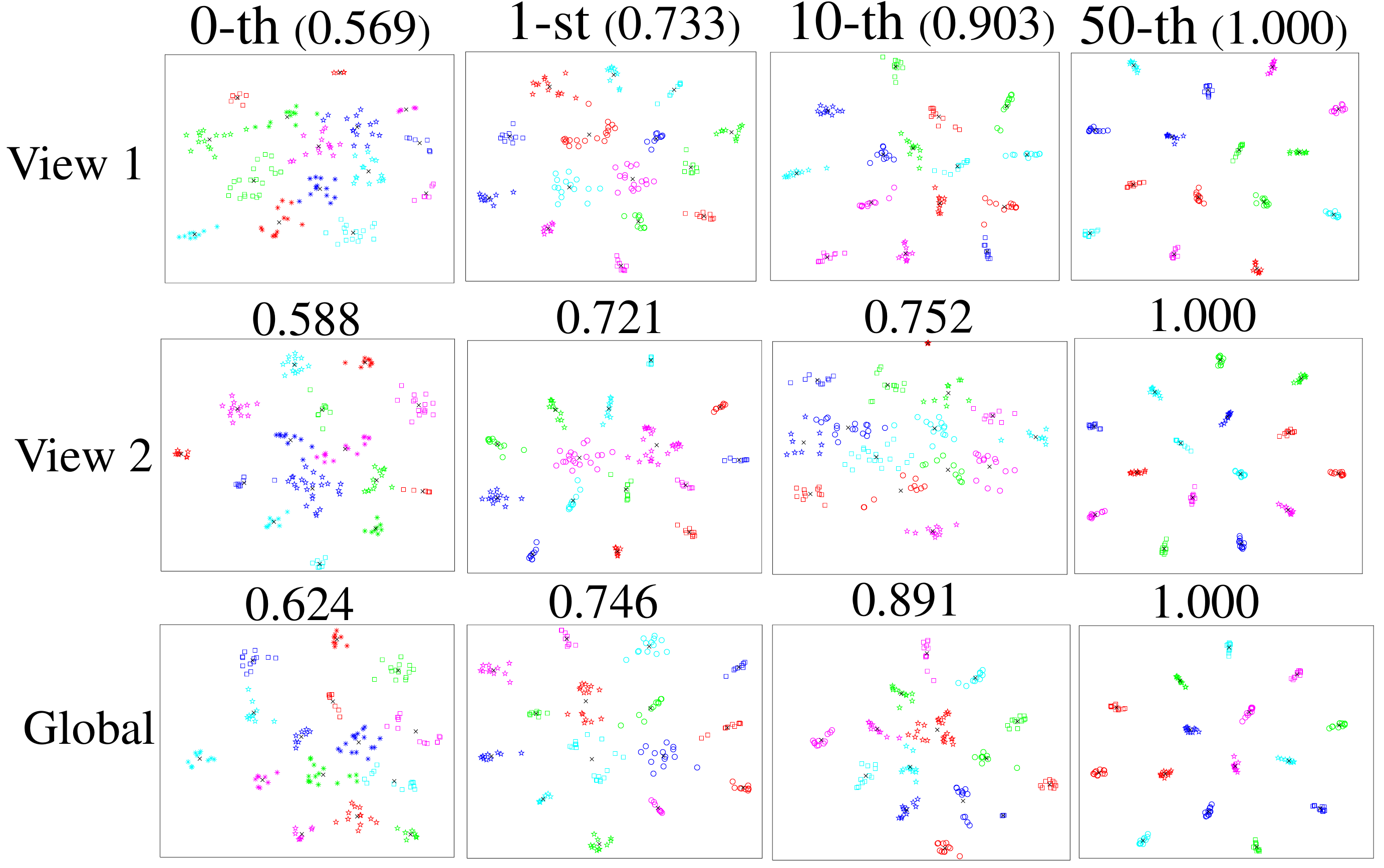}
\caption{ Visual illustration of latent representation v.s. learning iteration  on Yale data, 
where each feature is visualized by t-SNE~\cite{van2008visualizing}, and the value indicates the clustering accuracy (ACC) by K-means. 
 View 1 is intensity features; View 2 is  LBP features; Global is the fusion features from three different views. The four columns from left to right represent the clustering performance of the 0-th, 1-st, 10-th, and 50-th iterations, respectively.}
\label{fig:componet}
 \end{figure}
 
\subsection{Discussions}
\subsubsection{Model Analysis} Fig. \ref{fig:componet} shows how the clustering performance of the proposed MERA-MSC varies during the learning iteration on the Yale dataset. 
From the first column of Fig. \ref{fig:componet}, we can observe the clustering performance on all views is superior to that on a single view. Interestingly, when the low-rank MERA approximation is first used to explore inter/intra-view correlations, the clustering accuracy (ACC) increases by around 11\%, as illustrated in the second column of Fig. \ref{fig:componet}.
Iteratively, the correlation within views and across views mutually benefits from the low-rank MERA approximation. This allows us to learn a better latent structure from multi-view data, ultimately achieving a clustering accuracy of 100\%. Moreover, the performance improvement indicates that the utilization of information from both within and across views plays a significant role in enhancing clustering accuracy. Additionally, the low-rank MERA approximation effectively explores the inter-view and intra-view information in the self-representation tensor simultaneously.

\subsubsection{ Convergence Analysis}
\begin{figure}[!t]
 \centering
 \begin{subfigure}[b]{0.44\textwidth}
 \includegraphics[width=\textwidth]{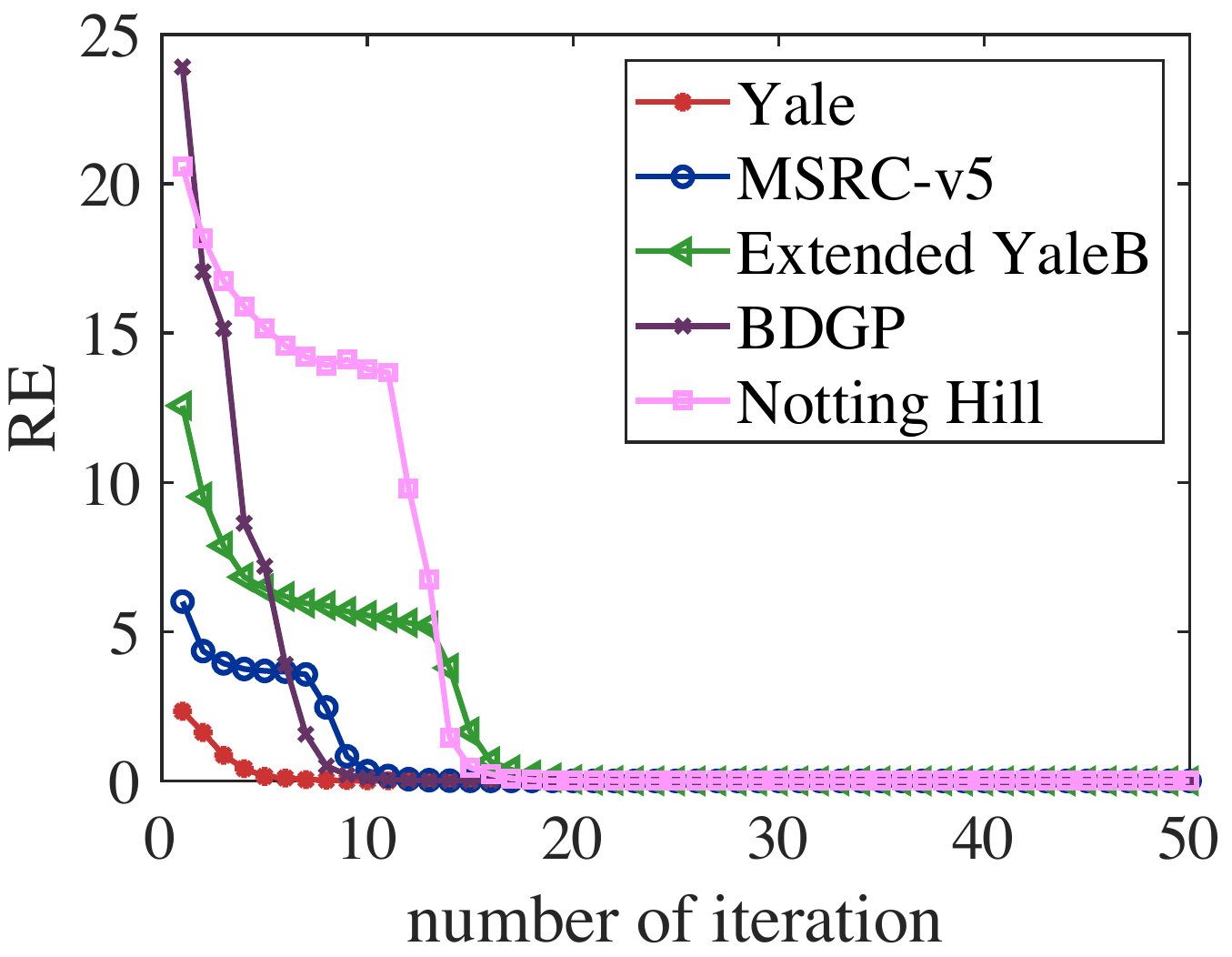}
    \caption{RE v.s. iteration}
    \label{fig:first11}
\end{subfigure}
\begin{subfigure}[b]{0.44\textwidth}
   \includegraphics[width=\textwidth]{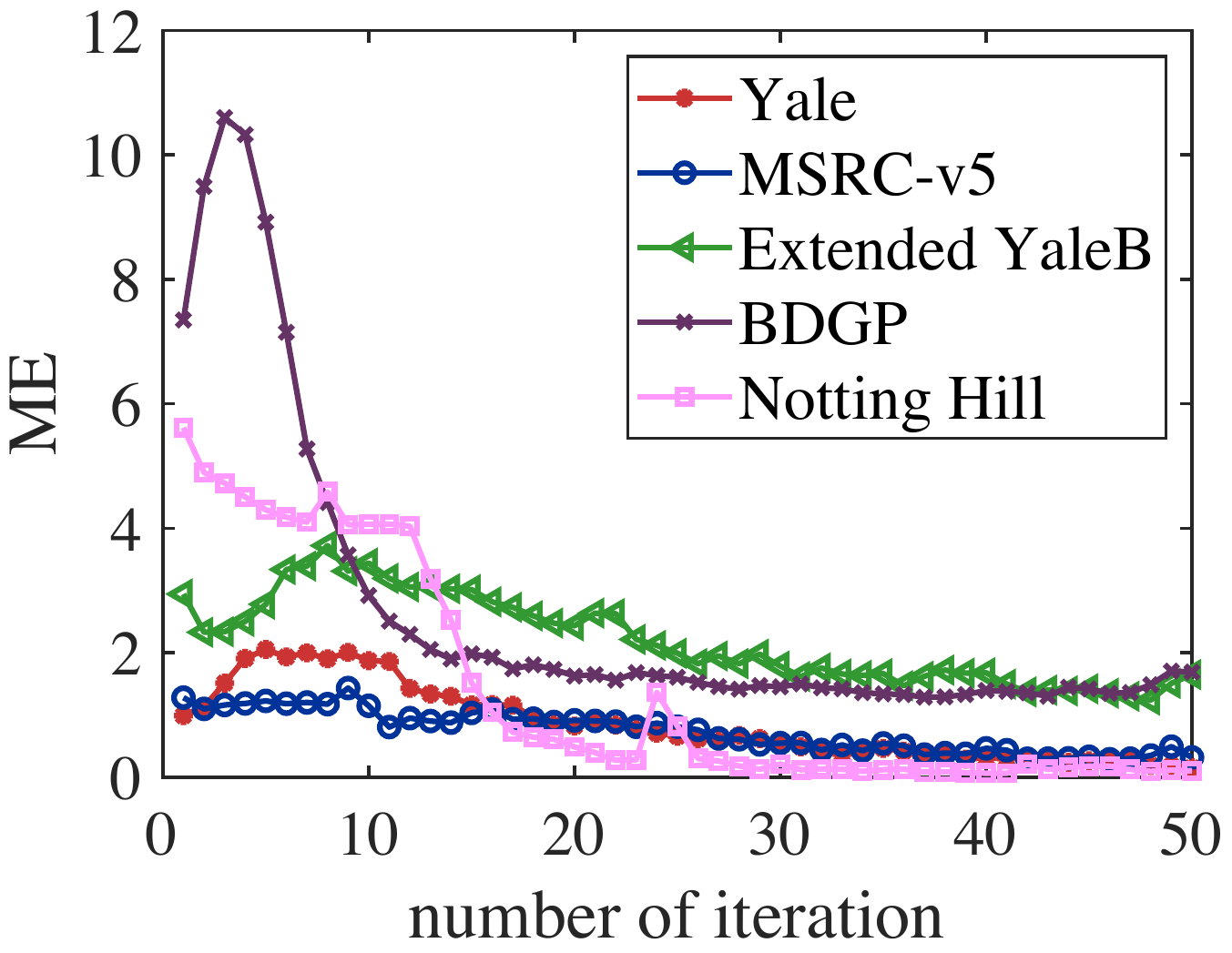}
    \caption{ME v.s. iteration }
    \label{fig:second11}
\end{subfigure}
\caption{The convergence performance of MERA-MSC.}
\label{fig:conv}
\vspace{-0.7cm}
 \end{figure}

Fig.~\ref{fig:conv} shows the convergence curves of MERA-MSC on different datasets, where the X-axis means the number of iterations and the Y-axis is the values of reconstruction error (RE) and match error (ME), which are defined by RE = $\max_{v,v = 1,\cdots, V}\|\mathbf{X}_v-\mathbf{X}_v\mathbf{Z}_v-\mathbf{E}_v\|_{\infty}$, and ME = $\max_{v,v = 1,\cdots, V}\|\mathbf{Z}_v-\mathbf{Y}_v\|_{\infty}$, respectively.
It can be observed the RE of MERA-MSC drops very fast at the first 10 iterations and eventually approaches 0 on all datasets.
In particular, the larger size of the datasets, the faster the RE and ME change. \\ 
\begin{table}[htbp]
    \centering
      \caption{Clustering results on CCV, Caltech-all and ALOI datasets.}
    \resizebox{\linewidth}{!}{
    \begin{tabular}{c|c|c|c|c|c|c|c}
\toprule \toprule \noalign{\smallskip}
Methods  &  F-score  &  Precision  &  Recall  &  NMI  &  ARI  &  ACC  & CPU Time (s)	\\ 
    \midrule 
 \multicolumn{8}{c}{ALOI ($R$=4, $\lambda$=0.02, $M$=10)}\\ 
\midrule 
 SMSC  &\textbf{0.7617}  &\textbf{0.7129}  &\underline{0.8185}  &\underline{0.8990}  &\textbf{0.7591}  &\textbf{0.8273}  & 
 365.7435  \\ 	
 LMVSC  &  0.3874  &  0.3498  &  0.4341  &  0.6964  &  0.3806  &  0.4897  &  93.0113  \\ 	
 SMVSC  &  0.2089  &  0.1366  &  0.4437  &  0.6033  &  0.1967  &  0.3251  &  423.8729  \\ 	
 FPMVS-CAG  &  0.1795  &  0.1053  &  0.6062  &  0.6334  &  0.1654  &  0.3121  &  180.5025  \\ 
 FastMICE  &  0.6713  &  \underline{0.6223}  &0.7289  &  0.8455  &0.6678  &  \underline{0.7543}  &\underline{20.3759}  \\ 
 sMERA-MVC  &\underline{0.7130}	 &0.6107	 &\textbf{0.8598}	 &\textbf{0.9114}	 &\underline{0.7096}	 &0.7424  &\textbf{19.5916} \\

\midrule 
 \multicolumn{8}{c}{Caltech-all ($R$=8, $\lambda$=0.1, $M$=9)}\\ 
\midrule 
 SMSC  &  0.1817  &  0.2935  &  0.1316  &\underline{0.4679}  &  0.1671  &  0.2448  &  269.1425 \\ 	
 LMVSC  &  0.1685  &  0.2810  &  0.1203  &  0.4374  &  0.1541	 & 0.2075  &92.7087  \\ 	
 SMVSC  &  0.1895  &  0.1200  &\textbf{0.4496}  &  0.3485  &  0.1516  &\underline{0.2906}  &	 356.5180  \\ 	
FPMVS-CAG  &\underline{0.2040}  &  0.1425  &\underline{0.3589}  &  0.3379  &\underline{0.1704}  &  0.2654  &  669.0376	  \\ 
FastMICE  &  0.1775  &\underline{0.3209}  &  0.1227  &  0.4370  &  0.1644  &  0.2090 &\textbf{34.8471}  \\ 	
 sMERA-MVC  &\textbf{0.3807}  &\textbf{0.5807}  &  0.2834  &\textbf{0.7533}  &\textbf{0.3690}	 &\textbf{0.4517}  &\underline{56.7210} 	 \\ 

\midrule 
 \multicolumn{8}{c}{CCV ($R$=2, $\lambda$=0.5, $M$=6)}\\ 
\midrule 
 SMSC  &\underline{0.1615}  &\underline{0.1599}  &\underline{0.1633}  &\underline{0.2184}  &\underline{0.1096}  &\underline{0.2714}  &  128.4686  \\ 	
 LMVSC  &  0.1202  &  0.1069  &  0.1374  &  0.1694  &  0.0592	 &  0.2073  &  59.2433  \\ 	
 SMVSC  &  0.1294  &  0.1235  &  0.1359  &  0.1540  &  0.0734  &  0.2110  &31.4677  \\ 	
FPMVS-CAG  &  0.1316  &  0.1174  &  0.1496  &  0.1597  &  0.0716  &  0.2282  &	 28.7811	  \\ 
FastMICE  &  0.1320  &  0.1387  &  0.1260  &  0.1795  &  0.0816  &  0.2306  &\underline{10.3551}  \\ 	
 sMERA-MVC  &\textbf{0.3603}	 &\textbf{0.3686}	 &\textbf{0.3526}	 &\textbf{0.5087}	 &\textbf{0.3221}	 &\textbf{0.4632}  &\textbf{6.0318} 	 \\ 

\midrule 
    \bottomrule
    \end{tabular}}
    \label{tab:large}
\end{table}
\subsubsection{Scalability analysis}
Most tensor-based MSC methods involve constructing an $N\times N$ relational graph with a computational complexity of $O(VN^3)$, where $N$ and $V$ represent the number of instances and views, respectively.
This process is time-consuming, particularly for large-scale multi-view data, as the computational time increases with the number of samples.


To extend  the MERA-MSC method, we have developed a scalable MERA based multi-view clustering (sMERA-MVC) algorithm. This approach replaces the self-representation learning step, denoted as $\mathbf{X}_v=\mathbf{X}_v\mathbf{Z}_v+\mathbf{E}_v$, in the MSC framework (equation (\ref{MERA model})), with anchor learning $\mathbf{X}_v=\mathbf{A}_v\mathbf{C}_v+\mathbf{E}_v$~\cite{sun2021scalable,wang2021fast}.
The revised framework is illustrated below:
\begin{equation}\label{MERA scale model}
\begin{aligned}
&\min_{\{\mathbf{E}_v,\mathbf{C}_v\}_{v=1}^{V}} \quad\sum_{v=1}^{V} \lambda\|\mathbf{E}_v\|_{2,1}\\
&\text{s.~t.~}  \mathbf{X}_v=\mathbf{A}_v\mathbf{C}_v+\mathbf{E}_v, \mathbf{A}_v^{\operatorname{T}}\mathbf{A}_v=\mathbf{I}_M, v=1,\cdots,V,\\
&\quad\hat{\mathcal{C}}=f(\mathbf{B},\mathcal{W},\mathcal{U}), \hat{\mathcal{C}}=\mathfrak{R}(\mathbf{C}_{1}, \mathbf{C}_{2}, \cdots, \mathbf{C}_{V}).
\end{aligned}
\end{equation}
Here, $\mathbf{A}_v\in\mathbb{R}^{D_v\times M}$ represents the anchor matrix, $M$ represents the number of anchors, and $\mathbf{C}_v\in\mathbb{R}^{M\times N}$ represents the anchor graph. The anchor graph captures the relationships between $N$ data points and $M$ anchors (where $M\ll N$) to represent the relationships among all data points, effectively reducing the computational complexity to $O(VNM^2)$.  $\hat{\mathcal{C}}=f(\mathbf{B},\mathcal{W},\mathcal{U})$ means that low-rank MERA approximation is considered on the anchor graph tensor to well explore the inter/intra-view correlations within multi-view data.

To validate the effectiveness and efficiency of sMERA-MVC, we conducted a comparative analysis with five fast MVC methods on three large-scale datasets. Below, we introduce the datasets used and outline the methods used for comparison.\\
\textbf{Datasets discription}:
 Amsterdam Library of Object Image (\textbf{ALOI})\footnote{http://elki.dbs.ifi.lmu.de/wiki/DataSets/MultiView}~\cite{geusebroek2005amsterdam} has a collection of 110250 images of 1000 small objects. Following the approach in~\cite{liang2022multi}, we use the first 100 classes as our testing dataset and obtained 10800 samples with four views.   Columbia Consumer Video (\textbf{CCV})\footnote{https://www.ee.columbia.edu/ln/dvmm/CCV/}~\cite{icmr11:consumervideo} is a rich database of YouTube videos containing 20 semantic categories. We remove the last 73 samples in our experiment, resulting in 6700 samples.
 \textbf{Caltech-all} \footnote{http://www.vision.caltech.edu/Image Datasets/Caltech101/}~\cite{fei2004learning} has 9144 samples and five views. In our experiment, we remove the last 24 samples to obtain 9120 samples.\\
\textbf{Methods discription}:
Sparse multi-view spectral clustering (SMSC) [2020, Neurocomputing]~\cite{hu2020multi}, 
large-scale multi-view subspace clustering (LMVSC) [2020, AAAI]~\cite{kang2020large}, scalable multi-view subspace clustering with unified anchors (SMVSC) 
[2021, ACM MM]~\cite{sun2021scalable}, fast Parameter-free multiview subspace clustering with consensus anchor guidance (FPMVS-CAG) [2021, TIP]~\cite{wang2021fast}, fast multi-view clustering via ensembles (FastMICE) [2023, TKDE]~\cite{huang2023fast}. \\
\textbf{Performance and Analysis on Large-Scale Multi-view Datasets}:
Table \ref{tab:large} shows the clustering performance of different methods on three large-scale multi-view datasets in terms of F-score, Precision, Recall, AR, NMI, ACC, and CPU time. 
We can observe that sMERA-MVC consistently outperforms the other five methods regarding clustering performance. In particular, for ALOI, SMSC performs well in most clustering accuracy metrics, with the sMERA-MVC method following closely behind.  Notably, sMERA-MVC achieves a significant advantage in terms of CPU time, running approximately 30 times faster than the SMSC.
For Caltech-all, sMERA-MVC demonstrates superior performance across all evaluation metrics, except for recall and computation time, surpassing all comparison baselines. Notably, sMERA-MVC excels in all aspects of the CCV dataset, while the performance of the other five methods is comparatively weaker. This further implies that the low-rank MERA approximation exhibits excellent performance in exploring the shared latent structure across views and within individual views in multi-view data.

\section{CONCLUSION}
In this paper, we propose a low-rank MERA based MSC algorithm, where the self-representation tensor can be adaptively learned from the multi-view data and low-rank MERA approximation in each iteration.
Benefiting from the interaction among orthogonal/semi-orthogonal factors of low-rank MERA approximation,  the correlations present across and within views can be effectively explored.
Numerical experiments on five well-known datasets show that MERA-MSC outperforms all state-of-the-art methods in clustering performance evaluated  on six different metrics. 
Besides, we have developed an accelerated and scalable MERA based multi-view clustering algorithm for large-scale multi-view data and verified its effectiveness and efficiency on three large-scale datasets. 






\bibliographystyle{plain}
\bibliography{cite}


\vfill

\end{document}